\documentclass[sigconf,screen,authorversion,nonacm]{acmart}

\AtBeginDocument{%
  }

\newcommand{\shortname}{{\scshape GenToC}}
\usepackage{myconstants}

\usepackage{multirow}
\usepackage{makecell}
\usepackage{cleveref}
\usepackage{algorithm}
\usepackage{algpseudocode}

\usepackage{pifont} %
\newcommand{\cmark}{\ding{51}}%
\newcommand{\xmark}{\ding{55}}%

\begin{document}

\title{A Framework for Leveraging Partially-Labeled Data for Product Attribute-Value Identification}

\author{D. Subhalingam}
\authornote{Both authors contributed equally to this research.}
\email{subhalingam.d@knowdis.ai}
\author{Keshav Kolluru}
\authornotemark[1]
\email{keshav.kolluru@knowdis.ai}
\affiliation{%
  \institution{KnowDis AI}
  \state{Delhi}
  \country{India}
}

\author{Mausam}
\email{mausam@cse.iitd.ac.in}
\affiliation{%
  \institution{Indian Institute of Technology, Delhi}
  \state{Delhi}
  \country{India}
}

\author{Saurabh Singal}
\email{saurabh@knowdis.ai}
\affiliation{%
  \institution{KnowDis AI}
  \state{Delhi}
  \country{India}
}

\begin{abstract}

In the e-commerce domain, the accurate extraction of attribute-value pairs (e.g., \textsc{Brand}: \textit{Apple}) from product titles and user search queries is crucial for enhancing search and recommendation systems.
A major challenge with neural models for this task is the lack of high-quality training data, as the annotations for attribute-value pairs in the available datasets are often incomplete.
To address this, we introduce \shortname{}, a model designed for training directly with partially-labeled data, eliminating the necessity for a fully annotated dataset.
\shortname{} employs a marker-augmented generative model to identify potential attributes, followed by a token classification model that determines the associated values for each attribute.
\shortname{} outperforms existing state-of-the-art models, exhibiting upto 56.3\% increase in the number of accurate extractions.
Furthermore, we utilize \shortname{} to regenerate the training dataset to expand attribute-value annotations.
This bootstrapping substantially improves the data quality for training other standard NER models, which are typically faster but less capable in handling partially-labeled data, enabling them to achieve comparable performance to \shortname{}.
Our results demonstrate \shortname{}'s unique ability to learn from a limited set of partially-labeled data and improve the training of more efficient models, advancing the automated extraction of attribute-value pairs.
Finally, our model has been successfully integrated into \href{http://www.indiamart.com}{IndiaMART}, India's largest B2B e-commerce platform, achieving a significant increase of 20.2\% in the number of correctly identified attribute-value pairs over the existing deployed system while achieving a high precision of 89.5\%.

\end{abstract}


\maketitle

\section{Introduction}
\label{sec:intro}

The rapid expansion of e-commerce has led to a significant increase in the variety and complexity of products available online. 
Each product typically includes a set of attributes such as \textsc{Brand}, \textsc{Model Name}, \textsc{Color}, with distinct values like \textit{Boat}, \textit{Rockerz 255 Pro}, \textit{Raging Red} (as demonstrated in \Cref{tab:example-query}).
These attributes and values help consumers locate and select their desired products.
Automatic attribute-value identification is a well-studied problem in the e-commerce literature 
\citep{bing2012unsupervised,putthividhya2011bootstrapped,probst2007semi,shinzato2013unsupervised,Roy2021AttributeVG,Wang2020LearningTE,Xu2019ScalingUO,Zheng2018OpenTagOA}.
It has been investigated in various contexts that involve additional metadata, such as product descriptions \citep{Yang2021MAVEAP,Shinzato2023AUG}, knowledge graphs \citep{ricatte2023avengr} or images \citep{Zhu2020MultimodalJA,Khandelwal2023LargeSG}. 
In this study, we focus on automated extraction using only textual information \citep{Xu2019ScalingUO,Zheng2018OpenTagOA}, such as product titles and user search queries.

To illustrate the practical utility of attribute-value extraction systems, we examine their usage in IndiaMART\footnote{\url{https://www.indiamart.com}}, India's largest B2B e-commerce platform, where our system is currently deployed.
In this platform, our attribute-value extraction system serves a dual purpose, enhancing both product listings and user searches.
The attribute-value pairs provided in product listings by sellers are often incomplete and can be enhanced using extractions from the product title.
In addition, the system is used to extract attribute-value pairs from user search queries, and these extracted pairs are matched with attribute-value pairs from the retrieved product listings to highlight the matching ones. 
This \textit{dynamic feature highlighting} of product attribute-value pairs based on the search query enables users to easily find the products that meet their requirements.

\begin{table}[b]
\centering
\caption{Complete collection of attribute-value pairs covering all words in the specified product title, with only the attributes \textsc{Brand}, \textsc{Model Name} and \textsc{Color} (marked with *) are included in the training data and the remaining attributes are extracted by our \shortname{} model.}
\label{tab:example-query}
\begin{tabular}{ll}
\multicolumn{2}{c}{\textbf{Product Title}: \textcolor{blue}{Boat} \textcolor{olive}{Rockerz 255 Pro}} \\
\multicolumn{2}{c}{\textcolor{red}{Raging Red} \textcolor{cyan}{Bluetooth} \textcolor{brown}{Neckband}} \\ \\ \toprule
\multicolumn{1}{l}{\textbf{Attribute}}                         & \multicolumn{1}{l}{\textbf{Value}}                            \\ \midrule
\textsc{Brand}*                                      & \textcolor{blue}{\textit{Boat}}                                      \\
\textsc{Model Name}*                          & \textcolor{olive}{\textit{Rockerz 255 Pro}}                           \\
\textsc{Color}*                                      & \textcolor{red}{\textit{Raging Red}}                                \\
\textsc{Connectivity}                               & \textcolor{cyan}{\textit{Bluetooth}}                                 \\
\textsc{Headphone Type}                             & \textcolor{brown}{\textit{Neckband}} \\ 
\bottomrule                  
\end{tabular}

\end{table}

The predominant approach for acquiring training datasets for attribute-value extraction tasks typically relies on leveraging product listings and associated attribute-value pairs as furnished by vendors on e-commerce platforms. 
This method is directly dependent on the comprehensiveness and accuracy of the data that vendors provide. 
However, it is common practice for vendors to list these details in an incomplete or inconsistent manner.
Such gaps and irregularities in the data present a substantial obstacle, as the development of effective deep learning models for attribute-value extraction is contingent upon the availability of large-scale, high-quality training data.
Given that attribute-value extraction systems are also employed on user query traffic to accurately capture user search intentions \citep{ricatte2023avengr}, it is essential to develop light-weight systems that provide real-time responses while managing the substantial workloads prevalent in e-commerce platforms.

\begin{figure}[t]
    \centering
    \includegraphics[scale=0.2]{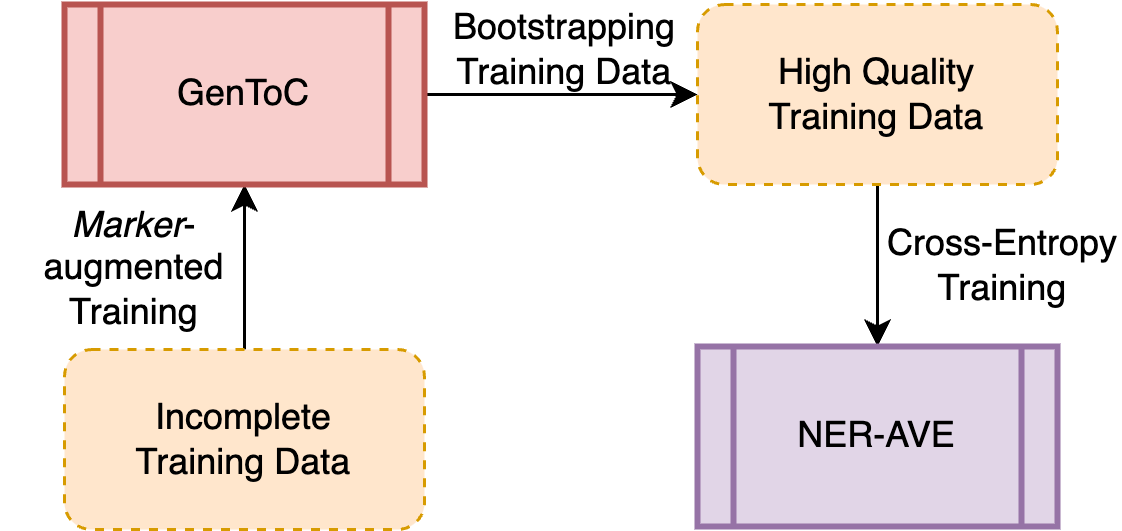}
    \caption{Overall framework. 
    We train \shortname{} system with markers to effectively learn from incomplete training data. It is then used to bootstrap high-quality training data to train the real-time NER attribute-value extraction (AVE) system.}
    \label{fig:pipeline}
\end{figure}

To address these dual challenges of incomplete training data and the need for real-time responses, we propose a novel framework.
First, we use a specialized model designed to learn from the incomplete data.
This model is then used to regenerate the training data, thereby enabling the training of a faster model that can deliver real-time responses.
This overall framework is illustrated in \Cref{fig:pipeline}.

To address the challenges of incomplete training data, we introduce \shortname{}, a novel model that uses a specialized \textit{marker} learning strategy.
\shortname{} is a two-stage neural model that decomposes the task into two distinct phases.
First, it identifies all attributes within the input text, which can be either a product title or a user search query.
Then, it extracts the corresponding values for each attribute.
Specifically, \shortname{} employs a \textbf{Gen}erative Seq2Seq model in the initial stage and a \textbf{To}ken \textbf{C}lassification model in the subsequent stage to accurately extract attribute-value pairs from the input text.
The first-stage generative model is designed to output all relevant attributes concatenated with each other using a delimiter.
The second-stage token classification model receives each attribute name along with the original input query and determines for each word in the text whether it belongs to the value of the input attribute.
For example, for the input text in \Cref{tab:example-query}, the first stage model can identify attributes such as \textsc{Brand}, \textsc{Model Name}, and \textsc{Color}. 
Then, the second stage model assigns \textit{Boat}, \textit{Rockerz 255 Pro} and \textit{Raging Red} as the values for the corresponding attributes.

To foster the model's learning capacity using partially-labeled data, we incorporate special \textit{markers} during the training phase of the first-stage generative model.
These markers are used to highlight the words in the input query that are present as attribute values in the training data.
For example, in \Cref{tab:example-query}, markers will be added to the words \textit{Boat}, \textit{Rockerz}, \textit{255}, \textit{Pro}, \textit{Raging}, \textit{Red} (at token level) because these are identified as attribute values in the training data.
Through the strategic use of these markers, the model learns to form associations between attributes and highlighted words, that every highlighted word is a potential source of attribute information.
This helps the model during inference, where we apply a marker to \emph{every} token in the input.
Applying a marker to each token serves as a signal to the model to regard every word as a possible source of attribute information, since similar correlations were observed during training. 
Thus, the model tends to produce a wider range of attributes and is not limited to mimicking the incomplete patterns observed in training.
Existing conventional models lack this generalization capability, making it difficult to utilize partially labeled data effectively.

The second-stage token classification model accepts both the attribute name and the original query as input.
It is tasked with classifying each word within the query to ascertain whether that signifies a value for the given input attribute. 
In the training phase, we use the attributes that are already known, whereas during inference, we rely on the attributes predicted by the first-stage model. 
Due to the cascading design of \shortname{}, errors from the first-stage model can potentially escalate, adversely affecting the second-stage model's performance. To fortify the second-stage model against such compounded inaccuracies, we implement a `Value Pruning' method. This technique equips the value-extraction model with the ability to generate null outputs for any incorrect attributes that might be output by the first-stage attribute-extraction model.

Compared to existing state-of-the-art NER models \citep{Zheng2018OpenTagOA} and Generative models \citep{Shinzato2023AUG} used for attribute-value extraction, \shortname{} achieves an increase of 16.2\% and 18.2\% in F1-score, respectively, on the test set. 
Despite these substantial performance gains, its speed of 90 ms per query presents a limitation for deployment.

Motivated by the objective of creating a faster model, we use \shortname{} to predict attribute-value pairs for each query in the training data and regenerate the dataset.
This bootstrapping process improves attribute-value tagging in the training dataset.
By training an NER model using this augmented dataset, we achieve an increase of 16.8\% in F1-score compared to using the original incomplete dataset.
It is also now comparable to \shortname{} in terms of F1-score, while maintaining an inference speed of under 9 ms per query.
The \shortname{}-bootstrapped NER system is currently deployed on IndiaMART, replacing a rule-based system and has already served over \TotalRequests{} requests. 
Compared to the previous system, our new system has improved the recall of attribute-value extraction by over 20\% on product titles and has resulted in a 9\% increase in user search queries that activate dynamic feature highlighting.
\looseness=-1

To summarize, the major contributions of our work include

\begin{itemize}
    
    \item Introducing \shortname{}, a novel two-stage architecture featuring a \textbf{Gen}erative Seq2Seq model for attribute extraction followed by a \textbf{To}ken \textbf{C}lassification model for value mapping, scalable to tens of thousands of attributes.
    \item Utilizing \textit{markers}-based learning in \shortname{}'s first stage to handle incomplete attribute-value tagging in training data.
    \item Surpassing existing methods with a 16.2\% F1-score increase.
    \item Enhancing the training data using bootstrapping, thereby boosting the F1-score of NER model trained with it by 16.8\%.
    \item Deploying the system in a leading B2B platform, where it has served over \TotalRequests{} requests in production so far.

\end{itemize}

\section{Related Work}
\label{sec:related_work}

Attribute-value extraction \citep{Nadeau2007ASO,bing2012unsupervised,putthividhya2011bootstrapped,probst2007semi,shinzato2013unsupervised,Roy2021AttributeVG,Wang2020LearningTE,Xu2019ScalingUO,Zheng2018OpenTagOA}
 has been a significant topic of study in the realm of e-commerce, with a wealth of research targeting the extraction of product details from various modalities, including purely text-based methods \citep{Wang2020LearningTE,Xu2019ScalingUO} as well as those that incorporate images \citep{Zhu2020MultimodalJA,Khandelwal2023LargeSG,Wang2023MPKGACMP}. 
A common limitation of these methods is their assumption of high-quality training data, which fails in many real-world scenarios where the training data is created using distant supervision and hence potentially incomplete \citep{Shrimal2022NERMQMRCFN,Xu2019ScalingUO}.
Some publicly available datasets like MAVE \citep{Yang2021MAVEAP} offer high-quality data (achieving over 98\% F1-score), but our work specifically addresses the more challenging case of partially-labeled settings, where existing models struggle.

Few works, such as \citet{Zhang2021QUEACOBT, Zheng2018OpenTagOA} focus on improving the training data quality in attribute-value extraction tasks but are severely limited to operating on a small number of attributes only.
For instance, \citet{Zhang2021QUEACOBT} rely on a subset of strongly-labeled data to train a teacher network which in turn creates training data for a student network.
The requirement of having a strongly-labeled subset limits them to operate on 13 attributes.
\citet{Zheng2018OpenTagOA} also relies on a small set of labeled instances to be used in an active learning setting to collect good examples for manual annotation.
They apply the technique to only one attribute per dataset.
In contrast, \shortname{} introduces a fundamentally new model design that can handle incomplete training data without requiring any completely labeled subsets.
This allows the model to scale to tens of thousands of attributes.

In terms of modelling, some of the earliest works on attribute-value extraction primarily employed rule-based extraction methods.
They utilized a specialized seed dictionary or vocabulary to identify key phrases and attributes \citep{Ghani2006TextMF,Wong2009ScalableAE,Gopalakrishnan2012MatchingPT}. 
Moving beyond rule-based systems, several neural models have also been proposed for this task in the recent past \citep{Roy2021AttributeVG}. 
Neural architectures for this task can be broadly divided into two categories -- token classification \citep{Yang2021MAVEAP,Xu2019ScalingUO,Shrimal2022NERMQMRCFN} or generative  \citep{Roy2022ExploringGM, Shinzato2023AUG} models.
Token classification models utilize NER models to identify the spans in the input text corresponding to an attribute.
On the other hand, generative models utilize Seq2Seq models to produce relevant attribute-value pairs from a specified input.
Our \shortname{} model makes clever use of both types of architectures, using a generative model for attribute extraction and a token-classification model for value extraction. 
We compare extensively with both types of architectures and show significant gains achieved by \shortname{}.

Token classification systems that rely on NER use atomic embeddings for encoding attributes, which makes it challenging to handle long-tail and complex attributes with less training data. 
Other token-classification systems utilize a BERT+LSTM model to embed each attribute separately \citep{Xu2019ScalingUO}.
The embedded representation attends over the input query and identifies the corresponding value. 
However, this approach encounters scalability issues as the model must check for values using every possible attribute. 
Some generative models \citep{Khandelwal2023LargeSG} are limited in their operation to a relatively small set of attributes (in their case, 38). 
This is because they use a Seq2Seq model pass for every attribute to identify the relevant values.
Other generative models \citep{Shinzato2023AUG} are applicable to larger attribute sets as they generate the attribute name as well.
Similarly, \shortname{} can handle large attribute sets due to a dedicated module that generates all the relevant attributes. 
Like other generative models, \shortname{} employs compositional encoding for attributes, which allows it to capture the semantic meaning of attribute names by considering the words they contain.
So it is better at handling long-tail and complex attributes compared to NER models. 
But the generative nature of the model comes with a higher response time.
We summarize the characteristics of the different models in \Cref{tab:architecture_comparison}.

\begin{table}[t]
\centering
\caption{Comparison of different types of models.}
\label{tab:architecture_comparison}
\begin{tabular}{lccc}
\toprule
Model & \makecell{Partially \\ labeled} & \makecell{Long-tail \\ attributes} & \makecell{Response \\ time} \\
\midrule
NER & \xmark & \xmark & fast \\
Seq2Seq & \xmark & \cmark & slow \\
\shortname{} & \cmark & \cmark & slow \\
\bottomrule
\end{tabular}
\end{table}

\begin{figure*}
    \centering
    \includegraphics[width=\textwidth]{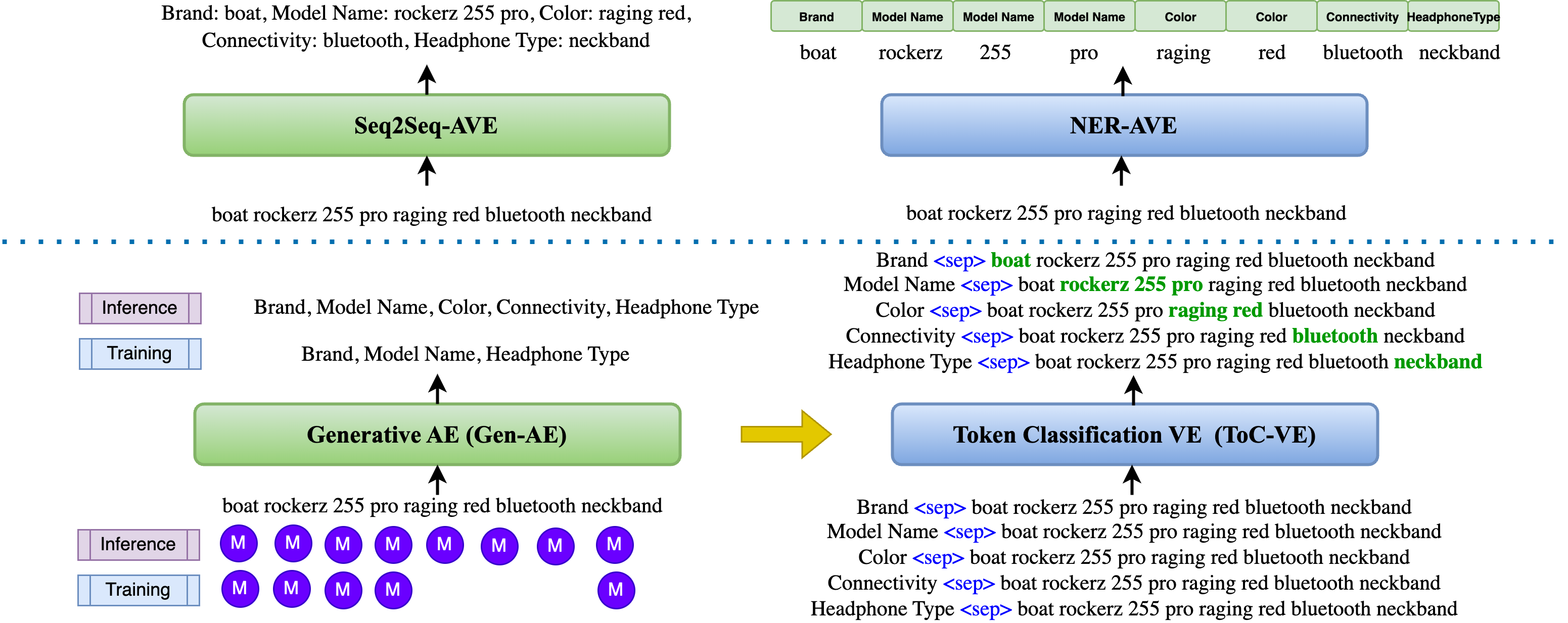}
    \caption{Model architectures. 
    (a) Seq2Seq-AVE outputs a string that concatenates all attribute-value pairs for a given input query. 
    (b) NER-AVE classifies each word in the query, tagging it with the relevant attribute.
    (c) \shortname{} employs Gen-AE to yield a concatenated list of attributes and ToC-VE to annotate the values linked to every recognized attribute. The Gen-AE model incorporates markers (`M') during the training process for the words which are covered. During inference, these markers are applied to all the words in the query.}
    \label{fig:gentoc}
\end{figure*}

\section{Problem Statement}
\label{sec:problem-statement}

Given a query $q$ with $k$ words, where $q = w_1 w_2 \dots w_k$, and a set of all potential attributes $\mathbb{A}$, the objective of attribute-value extraction is to identify all possible attribute-value pairs, denoted as $\{(a_1, v_1), (a_2, v_2), \dots, (a_n, v_n)\}$, where $a_i \in \mathbb{A}$ and $v_i$ is a subset of words in $q$.
In our setup, the input $q$ may be a product title or user search query, with each word $w$ in $q$ linked to at most one attribute.

\section{Background}
\label{sec:background}

In this section, we describe in detail two popular architectures that have been commonly used for the task of attribute-value extraction.
We illustrate their working in \Cref{fig:gentoc} (a) and (b).

\subsection{NER-AVE}

Following NER models used for the task \citep{Rezk2019AccuratePA,Shinzato2022SimpleAE}, NER-AVE is an encoder-only model that operates by classifying tokens in the input query, where each token in the query is assigned a label corresponding to the relevant attribute. 
For example, \textit{boat} is assigned the attribute \textsc{Brand}, each of three words \textit{rockerz}, \textit{255} and \textit{pro} are assigned the attribute \textsc{Model Name}, and so on.
In case no attribute exists for a particular word, a special label, \textsc{NoAttribute} is used to indicate the same.
As an NER model, it treats every attribute as a unique label and assigns an atomic embedding. 
However, its major strength lies in its speed due to the simple encoder-only architecture. 
Although it is not capable of effectively learning from partially labeled data, we find that training it with better quality data generated by \shortname{}, results in a fast and powerful attribute-value extraction system.

\subsection{Seq2Seq-AVE}

The Seq2Seq-AVE model is based on the model developed in \citet{Shinzato2023AUG}, which employs a Seq2Seq model that yields a concatenated string, incorporating the respective attribute-value pairs, for a given input query.
For instance, the encoder takes an input $q$, and the decoder generates the output string `$a_1$:$v_1$, $a_2$:$v_2$, $\dots$, $a_n$:$v_n$'.
For example, the decoder produces the generation ``\textit{Brand: boat, Model Name: rockerz 255 pro ...}'', one token at a time.
The attribute and value pairs can then be parsed from this generation.
Its limitations are slow inference speed (found to be over 10x slower than NER-AVE) and limited learning ability from partially labeled data.

\section{Methods}
\label{sec:methods}

In this section, we describe the two-stage \shortname{} model and the marker-based training that enables it to learn from partially-labeled data.
We show the working of \shortname{} in \Cref{fig:gentoc} (c).
The initial step employs a Generative Attribute Extraction (Gen-AE) model, which takes the product title or user search query as input and subsequently generates a concatenated list of predicted attributes.
The model is trained to generate attributes in the order their values occur in the input.
Given the model's generative ability, it might generate attributes outside the training set. However, these are very infrequent (occur less than 0.1\% of the times), so we forgo constrained generation \citep{decao2021autoregressive}, avoiding any restrictions that would force the attributes to belong solely to the original set of attributes.

\paragraph{Markers:}
To deal with partially-labeled training data, we initially identify the words within the input query that correspond to values of any attribute. We refer to these identified words as \textit{marked words}.
For the example, ``\textit{Boat Rockerz 255 Pro Raging Red Bluetooth Neckband}'', in \Cref{tab:example-query}, [\textit{boat}, \textit{rockerz}, \textit{255}, \textit{pro}, \textit{raging}, \textit{red}] are treated as marked words during training, as they have been labeled with some attribute. 
A special learnable embedding, termed as ``\textit{marker embedding}'', is added to the encoder's final hidden states of every token of the marked words before being passed to the decoder.
The same learnable embedding is shared across all marked tokens everywhere.
The model is then able to learn that the output attributes, as observed in the training data, are a result of considering only the limited set of words in the input that have been marked. 
In other words, the marker embeddings act as signals that instruct the model to focus on and incorporate the information from these marked words into its attribute generation process, while preserving the broader context of the input query.

At inference time, the marker embedding is applied to all words within the input query, nudging the model to generalize and output attributes that are relevant to any word present in the query.
This enhances the model's capability to recognize and associate attributes with words like \textit{bluetooth} and \textit{neckband}, which did not have valid attribute-value pairs annotated and hence were not marked at training time.

In the next stage, a Token Classification Value Extraction (ToC-VE) model takes each of the attribute names from the first stage along with the original query, separated by a special delimiter, \textit{<sep>}.
It then labels each token with a binary value, for \textit{yes} or \textit{no}, to denote whether it corresponds to a value or not for the chosen attribute.
This model is trained independently of the first-stage model.
Since attribute \emph{names} are used in the training process, it allows the model to learn from closely related attributes which share similarities in their names.
This becomes particularly crucial when dealing with a noisy attribute name ontology, which may include different attribute names conveying the same property, such as \textsc{Model Number} and \textsc{Model No}.

\paragraph{Value Pruning:} 
When training the ToC-VE model with a set of attribute-value pairs, it always contains a value for every attribute. 
However, during the inference phase, the attributes may be erroneously generated by the Gen-AE model. 
As a consequence, ToC-VE model would tend to assign some value to even the incorrect attributes.
To counteract this, we supplement the ToC-VE training with additional data to identify instances where no correct value is present for a given attribute. 
We accomplish this by taking an existing attribute-value pair and deleting the value from the original query. 
Consequently, we ensure that the chosen attribute no longer appears in the query, training the model to label all tokens as `NO' values for the attribute.
For example, if we remove the token, `\textit{boat}', the example, ``\textit{brand \textit{<sep>} rockerz 255 pro raging red bluetooth neckband}'', will have no value for the attribute \textsc{brand}.
We term the generation of this kind of synthetic training data as Value Pruning.
For the final set of attribute-value pairs yielded by the model, we exclude those attributes that do not have any associated values.

Therefore, the architecture of \shortname{} ensures effective attribute-value pair extraction, preventing error build-up in the pipeline, even in the face of partially labeled data and vast attribute sets that contain redundancies.
The complete algorithm for training and inference are described in \Cref{algo:training}  and \Cref{algo:inference} (in Appendix), respectively.

\section{Experimental Setup}
\label{sec:setup}

\subsection{Dataset and Metrics}
\label{sec:dataset-and-metrics}

\begin{figure}
    \centering
    \includegraphics[scale=0.5]{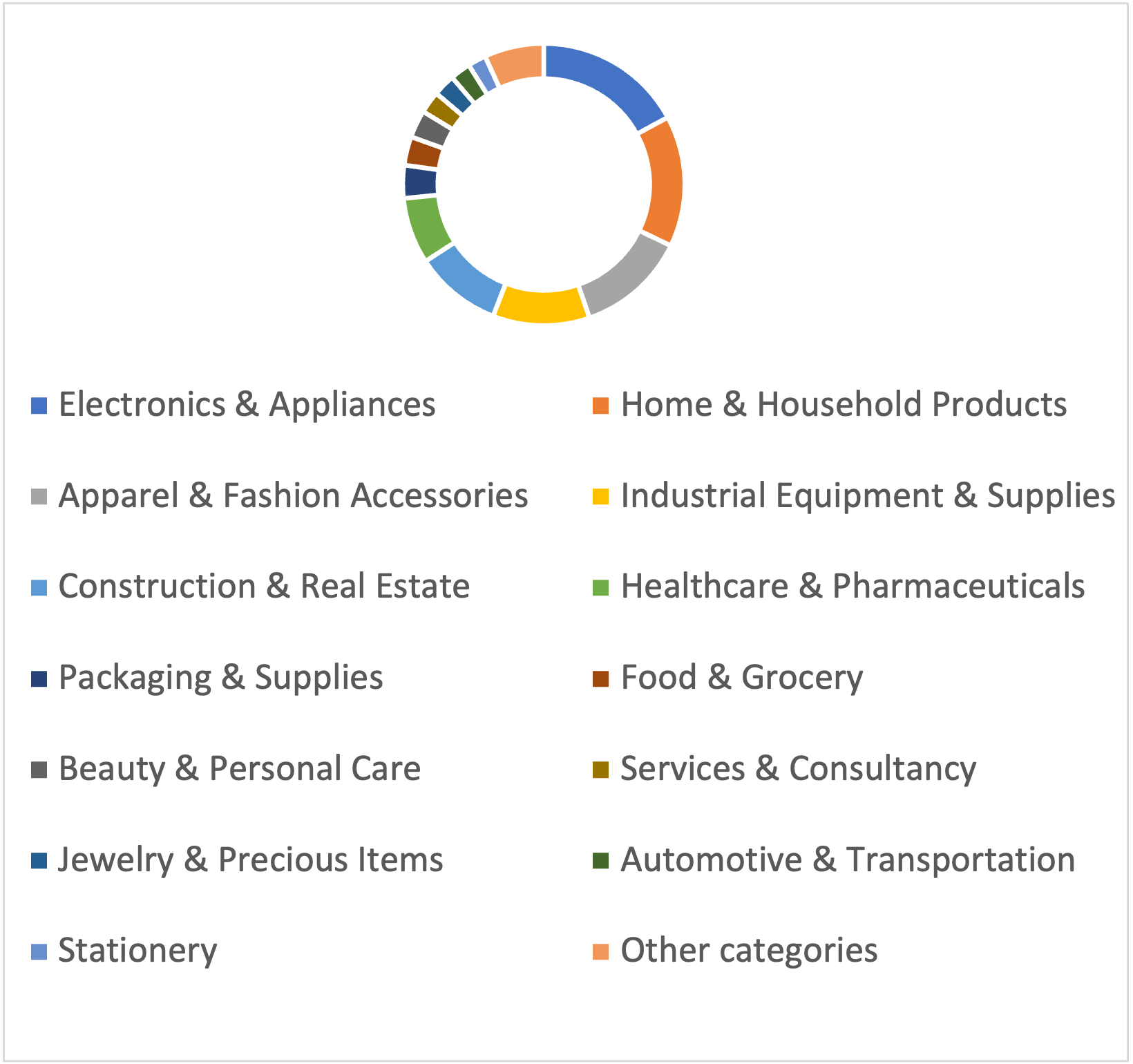}
    \caption{
    Distribution of product categories within training dataset. 
    Specific percentage values are omitted to preserve data confidentiality.}
    \label{fig:dist-chart}
\end{figure}

To curate data for the attribute-value extraction task, we make use of the product specifications that are provided by the sellers for product listings on IndiaMART.
IndiaMART has \NumBuyers{} buyers, \NumSellers{} sellers, and features over \NumProducts{} different products and services.
For the experiments in this paper, we limit our input exclusively to product titles and use a set of \NumTrainingSamples{} examples for training models.
It comprises \NumUniqueAVPairs{} unique attribute-value pairs covering \NumAttr{} attributes.
The category-wise distribution of products is presented in \Cref{fig:dist-chart}.
\Cref{tab:top-5-attributes} lists the top-five attributes and provides a set of example values for each.
Despite the large scale of attributes, only 40.7\% of the words of a product title are tagged on average with an attribute.
This motivates us to build models designed for partially labeled data, where the present attribute-value pairs are reliable and of high quality but might lack the complete set of attribute-values.

\begin{table}
    \small
    \centering
    \caption{Top-5 attributes in the dataset (based on frequency), along with a set of example values for each.}
    \label{tab:top-5-attributes}
    \begin{tabular}{@{}ll@{}}
        \toprule
        \textbf{Attribute} & \textbf{Values} \\
         \midrule
         \textsc{Brand}              & \textit{hp}, \textit{samsung}, \textit{dell}, \textit{siemens}, \textit{bosch} \\
         \textsc{Material}           & \textit{stainless steel}, \textit{plastic}, \textit{brass}, \textit{wooden}, \textit{cotton}\\
         \textsc{Color}              & \textit{black}, \textit{white}, \textit{blue}, \textit{red}, \textit{brown} \\
         \textsc{Model name/number}  & \textit{n95}, \textit{classic}, \textit{12a}, \textit{kn95}, \textit{eco} \\
         \textsc{Usage}              & \textit{industrial}, \textit{office}, \textit{kitchen}, \textit{packaging}, \textit{home} \\
         \bottomrule
    \end{tabular}
\end{table}

We evaluate the systems on \NumAutoTestingSamples{} samples (\textbf{\NameAutoTesting{}}) based on the ground truth values available. 
We compute the precision, recall and F1-score by comparing the set of attribute-value pairs generated for each input with the ground truth set of attribute-value pairs.
We then report their averages taken across all examples.
However, we find that automatic evaluation is challenging and often unreliable due to the lack of normalization in the attribute names, as the ground truth set can use different attributes to express the same characteristic. 
So, we randomly sample \NumManualTestingSamples{} examples (\textbf{\NameManualTesting{}}) and get every output checked using 3 data annotators (DAs). These annotators are skilled professionals who perform various annotation tasks within our organization.
We consider the majority class assigned by the DAs to determine if a given attribute-value pair is correct/incorrect.
We find that 
the inter-annotator agreement, computed using Fleiss' kappa \citep{Fleiss1971MeasuringNS}, is $\kappa = \InterAnnotatorAgreementKappa{}$, indicating substantial agreement among the annotators \citep{Landis1977}.
The response time is measured by calculating the average time taken for queries from the 2K test set (with a batch size of one) using an NVIDIA GeForce RTX 3090 GPU.

\subsection{Implementation and Systems Compared}
\label{sec:baselines-and-implementation}

To compare \shortname{} with the two classes of state-of-art models, we use NER-AVE as a representative of the available NER-style models \citep{Zheng2018OpenTagOA,Rezk2019AccuratePA,Shinzato2022SimpleAE} and Seq2Seq-AVE as a representative of the published generative models \citep{Roy2022ExploringGM,Shinzato2023AUG}.
We use our own implementations due to the lack of availability of standard open-source implementations for the above works.
For a fair comparison, we use the DeBERTa-V3-Small\footnote{\url{https://hf.co/microsoft/deberta-v3-small}} \citep{debertav3} for all the classification tasks and BART-Base\footnote{\url{https://hf.co/facebook/bart-base}} \citep{mbart} for all generation tasks. Both architectures consist of 6 encoder layers, and BART has 6 additional decoder layers. 
Due to the need for industry-scale systems to be used by millions of users, we don't experiment with LLMs, which are highly resource intensive.

\section{Experiments and Results}
\label{sec:expts}
In this section, we answer the following questions:

\begin{enumerate}
    \itemsep0em 
    \item How does \shortname{} compare to other models using both automated and manual evaluation?   %
    \item What is the incremental contribution of Markers and Value Pruning in the \shortname{} system? %
    \item Can \shortname{} improve training data tagging and thus be used to train faster models? %
    \item What are the trade-offs between precision and recall in various models at different confidence thresholds?
    \item How is the system utilized and how does it perform in a production environment?
\end{enumerate}

\begin{table*}
\centering
\caption{Model Performance.
A comparison of NER-AVE, Seq2Seq-AVE and \shortname{} models reveals that \shortname{} performs superiorly in both automated and manual evaluations in terms of F1-score.
However, it comes at the cost of increased response time with respect to NER-AVE and thus motivates our bootstrapping experiments.}
\label{tab:performance-main}
\begin{tabular}{@{}lrrrrrrr@{}}
\toprule
\multirow{2}{*}{\textbf{Architecture}} & \multicolumn{3}{c}{\textbf{Automatic Evaluation (39K)}}                   & \multicolumn{3}{c}{\textbf{Manual Evaluation (2K)}} & \multirow{2}{*}{\textbf{\begin{tabular}[c]{@{}r@{}}Response Time\\ (in ms/query)\end{tabular}}} \\ \cmidrule(l){2-4} \cmidrule(l){5-7} 
                                    & \textbf{Precision} &  \textbf{Recall} & \textbf{F1-score} & \textbf{Precision}  & \textbf{Recall} & \textbf{F1-score} &                                                                                                    \\ \midrule
NER-AVE                                           & 80.6    & 60.1            & 68.9      & 90.8   & 52.8 & 66.8          & \textbf{8.8 \tiny{± 1.1}}                                                                                                 \\
\ \ \ \ \ \ \ \ with Marker                                        & 58.6     & 71.2          & 64.3        &         66.5      & \textbf{87.5} & 75.6                  & \textbf{8.9 \tiny{± 1.1}}                                                                                                 \\ \midrule
Seq2Seq-AVE                                           & 80.1     & 52.5           & 63.4          & 89.6    & 50.7 & 64.8                    & 91.9 \tiny{± 23.8}                                                                                               \\
\ \ \ \ \ \ \ \ with Marker                                       & 60.0      & 67.9          & 63.7               &  65.2    & 69.9 &  67.5                    & 149.4 \tiny{± 43.5}                                                                                             \\ \midrule
\shortname{}                                         & 70.8    & 71.8               & \textbf{71.3}               & 86.1     & 80.1 & \textbf{83.0}                   & 90.1 \tiny{± 19.6}                                                                                            \\
\ \ \ \ \ \ \ \ without Marker                                   & \textbf{80.8}    & 54.7             & 65.2                      & \textbf{92.7}  & 51.0 & 65.8                      & 66.1 \tiny{± 12.6}                                                                                               \\
\ \ \ \ \ \ \ \ without VP                                        & 66.3    & \textbf{74.3}            & 70.1                      & 78.9    & 85.4 & 82.0                    & 86.3 \tiny{± 18.8}                                                                                               \\
\ \ \ \ \ \ \ \ without Marker \& VP                          & 79.7  & 55.1                 & 65.2                      & 90.7    & 52.2 & 66.3             & 66.5 \tiny{± 12.2}                                                                                              \\ \bottomrule \\
\end{tabular}
\vspace{1em}
\end{table*}

\begin{table*}
\centering
\caption{Bootstrapping performance. \shortname{} and NER-AVE are re-trained with data bootstrapped from the two models. Results for \shortname{} and NER-AVE trained on original data are included for reference. NER-AVE trained with data bootstrapped from \shortname{} gives us the best trade-off between performance and speed for deployability.}
\label{tab:performance-bootstrap}
\begin{tabular}{@{}llrrrrrr@{}}
\toprule
\multirow{2}{*}{\textbf{Data Source}} & \multirow{2}{*}{\textbf{Architecture}} & \multicolumn{3}{c}{\textbf{Automatic Evaluation (39K)}}                   & \multicolumn{3}{c}{\textbf{Manual Evaluation (2K)}} \\ \cmidrule(l){3-5}\cmidrule(l){6-8} 
                                       &                                       & \textbf{Precision}  & \textbf{Recall} & \textbf{F1-score} & \textbf{Precision}  & \textbf{Recall} & \textbf{F1-score}      \\ \midrule
Original & NER-AVE                                           & \textbf{80.6}    & 60.1            & 68.9      & \textbf{90.8}   & 52.8 & 66.8                              \\
& \shortname{}                                         & 70.8    & 71.8               & 71.3               & 86.1     & 80.1 & 83.0                   \\ \midrule                              
NER-AVE                                  & NER-AVE                                &  80.1     & 61.3                 & 69.5              & 90.0        & 55.3 & 68.5                                 \\
& \shortname{}                                 & 70.7  & 68.8                  & 69.7              & 86.8  & 76.8 & 81.5        \\ \midrule
\shortname{}                                      & NER-AVE                                &  70.9   & \textbf{72.4}             & \textbf{71.6}      & 85.6     & 81.6 &    83.6                                  \\
& \shortname{}                                 & 64.5   & \textbf{72.4}             & 68.2              & 83.2  & \textbf{91.2} & \textbf{87.0}                                        \\ \bottomrule \\
\end{tabular}
\vspace{1em}
\end{table*}

\begin{table*}
\centering
\caption{Predictions made by NER-AVE, Seq2Seq-AVE, and \shortname{} for three test examples. \shortname{} correctly identifies more attribute-value pairs in comparison to NER-AVE and Seq2Seq-AVE. (\textit{SS} stands for \textit{stainless steel})}

\label{tab:example-predictions}
\begin{tabular}{cccc}
\toprule
\textbf{Product Title} & \textbf{NER-AVE} & \textbf{Seq2Seq-AVE} & \textbf{\shortname{}} \\
\midrule
\makecell[c]{Casual Juliet Sleeve Solid \\ Women Maroon Top} & \makecell{\textsc{Color}: \textit{Maroon}} & \makecell{\textsc{Color}: \textit{Maroon}} & \makecell{\textsc{Occasion}: \textit{Casual} \\ \textsc{Sleeves Type}: \textit{Juliet Sleeve} \\ \textsc{Pattern}: \textit{Solid} \\ \textsc{Gender}: \textit{Women} \\ \textsc{Color}: \textit{Maroon}} \\
\midrule
\makecell{Sofar Ongrid Inverter \\ 5.5KTL-X 5.5KW Three Phase} & \makecell{\textsc{Brand}: \textit{Sofar}} & \makecell{\textsc{Brand}: \textit{Sofar} \\ \textsc{Model}: \textit{5.5KTL-X}} & \makecell{\textsc{Brand}: \textit{Sofar} \\ \textsc{Grid Type}: \textit{Ongrid} \\ \textsc{Model}: \textit{5.5KTL-X} \\ \textsc{Capacity}: \textit{5.5KW} \\ \textsc{Phase}: \textit{Three Phase}} \\
\midrule
\makecell{Globe SS Induction \\ Pressure Cooker} & \makecell{\textsc{Brand}: \textit{Globe}} & \makecell{\textsc{Brand}: \textit{Globe}} & \makecell{\textsc{Brand}: \textit{Globe} \\ \textsc{Material}: \textit{SS} \\ \textsc{Type of Pressure Cookers}: \textit{Induction}} \\
\bottomrule \\
\end{tabular}
\end{table*}

\subsection{Comparison with Other Models}
\label{sec:baseline-comparison}

In \Cref{tab:performance-main}, we present a comparison of three systems -- NER-AVE, Seq2Seq-AVE and \shortname{}.
\shortname{} model outperforms its counterparts in recall, achieving 71.8\% in automatic and 80.1\% in manual evaluations, which are the highest among the three models.
These scores are notably higher (by 11.7\% and 27.3\% in automatic and manual evaluations) than those of the next best-performing model, NER-AVE.
Such a significant gap in recall demonstrates that \shortname{} effectively leverages the partially labeled nature of the training data, resulting in more extractions.
Indeed, the total number of correct attribute-value extractions rose from 2,636 to 4,121 (a 56.3\% increase) on the 2K test set.
This increase in recall is mainly due to Markers, which is elaborated upon in \Cref{sec:ablation-study}.
In terms of precision, both the NER-AVE and Seq2Seq-AVE models exhibit higher precision than the \shortname{} model in both automatic and manual evaluations, with approximately 9-10\% and 4-5\% higher precision, respectively. 
Overall, \shortname{} maintains its superiority in F1-score, recording the highest at 71.3\% in automatic evaluations and 83.0\% in manual evaluations, indicating a more balanced and consistent performance.
However, a drawback of \shortname{} is its slower response time compared to the NER-AVE model. We address a potential solution to this performance-speed trade-off in \Cref{sec:bootstrapping}.

In \Cref{tab:example-predictions}, we present example cases showcasing \shortname{}'s superior performance in comparison to both NER-AVE and Seq2Seq-AVE models, in terms of the number of attribute-value pairs extracted.

NER-AVE and Seq2Seq-AVE models are unable to leverage the partially-labeled nature of the training data, as indicated previously in \Cref{tab:architecture_comparison}, despite possibly encountering the remaining unlabeled attribute-value pairs in other examples.
In particular, each training example is tagged with only 1.53 attribute-value pairs on average, so it's not unexpected that these models maintain a similar tagging frequency during inference.
\shortname{} addresses this challenge by incorporating Markers, which allow for the identification of more number of attribute names than what is typically found in the training data.

For instance, in the product title ``\textit{Casual Juliet Sleeve Solid Women Maroon Top}'' from the first example in \Cref{tab:example-predictions}, the \shortname{} model surpasses the other models by identifying four additional attributes -- \textsc{Occasion}, \textsc{Sleeves Type}, \textsc{Pattern} and \textsc{Gender}.

\subsection{Ablation Study}
\label{sec:ablation-study}

Our initial hypothesis posited that incorporating Markers would enhance recall by identifying more attributes (critical in dealing with partially-labeled data), while Value Pruning (VP) would boost precision by eliminating inaccurately generated attributes. 
To validate this, we evaluated variations of the \shortname{} model, specifically configurations excluding either Markers, VP, or both.
The outcomes, as presented in \Cref{tab:performance-main}, align with our expectations.
Notably, in the 2K test set, the absence of Markers in the \shortname{} model resulted in a 29.1\% decrease in recall, while omitting VP led to a 7.2\% reduction in precision. This pattern was also evident in the larger 39K test set.
The findings clearly demonstrate the critical roles of both Markers and VP in balancing recall and precision.
The \shortname{} model, when devoid of either component, shows a diminished F1-score compared to its complete configuration.
These results substantiate our choice to incorporate both these strategies in the final formulation of the \shortname{} model, for its superior overall performance.

We also experiment with adding Markers to NER-AVE and Seq2Seq-AVE.
During training, we incorporate marker embeddings to those words which have been covered by the available attributes, and at inference, we incorporate the marker embeddings to all words. Consistent with earlier practices, these embeddings are added into the final hidden states of the encoder.
In the 2K test set, this modification results in considerable reductions in precision scores, plummeting to 66.5\% from 90.8\% in NER-AVE and to 65.2\% from 89.6\% in Seq2Seq-AVE. A similar pattern can be observed in case of the 39K test set as well. 
This outcome underscores the importance of two-stage model for using markers effectively.
The bifurcated structure of the two-stage model ensures that markers are exclusively utilized in the first stage. 
This design circumvents the problem in single-stage models, where markers can force the assignment of attributes to every word, often resulting in suboptimal performance.
This also significantly increases the response time for the Seq2Seq-AVE model (see \Cref{tab:performance-main}), as it has to generate a longer output due to more extractions.
On the other hand, Value Pruning is useful when there is an candidate set of attribute names that might include some incorrect ones, such as those generated by the first stage of the \shortname{} model.
It aims to reduce irrelevant attribute names in the final result by predicting empty values for them.
Conversely, Seq2Seq-AVE and NER-AVE produce both attribute-value pairs in a unified step, making Value Pruning inapplicable for them.

\subsection{Bootstrapping Training Data}
\label{sec:bootstrapping}

From \Cref{tab:performance-main}, we find that NER-AVE has the fastest response time.
Seq2Seq-AVE is the slowest, as it has to generate a long string containing all attributes and values.
The high performance of \shortname{} comes at the cost of speed, as its response time is 10x greater than NER-AVE's.
With the motivation to build a deployable system that can handle real-world traffic, we experiment with cleaning the training data by regenerating it using the trained \shortname{} model.
It increases the average number of words tagged in any attribute-value pair from 0.41 to 0.65 while increasing the total number of attribute-value pairs from 3M to 4.7M on the full training data.
In \Cref{tab:performance-bootstrap}, we find that the NER-AVE model trained with this \shortname{}-bootstrapped training data significantly improves the performance over the NER-AVE model trained with original partially-labeled data.
In fact, its performance when trained with \shortname{}-bootstrapped data is comparable to the \shortname{} model in both automatic and manual evaluations.
This makes it a strong alternative to the \shortname{} model for production environments, where rapid response times are crucial.
Additionally, we experiment with regenerating the training data using the originally trained NER-AVE, the second-best model in terms of F1-score.
However, we find that training with this data results in marginally decreased performance, proving less effective than training with \shortname{}-bootstrapped data.
This is because, unlike \shortname{}, the NER-AVE model is not equipped to learn from incomplete data. Consequently, it does not help in reducing the partially labeled nature of the training data upon regeneration, rendering the bootstrapping unproductive.
Furthermore, in \Cref{sec:performance-long,sec:performance-long-tail-attributes}, we also assess the performance of these models on long product titles and long-tail attributes. We find that \shortname{} outperforms Seq2Seq-AVE and NER-AVE systems by more than 13-18\% in F1-score for long product titles and 7-19\% in precision for rare attributes while helping boostrapped NER-AVE achieve similar levels.

\subsection{Precision-Recall Trade-off}
\label{sec:pr-curve}

\begin{figure}
    \centering
    \includegraphics[width=\columnwidth]{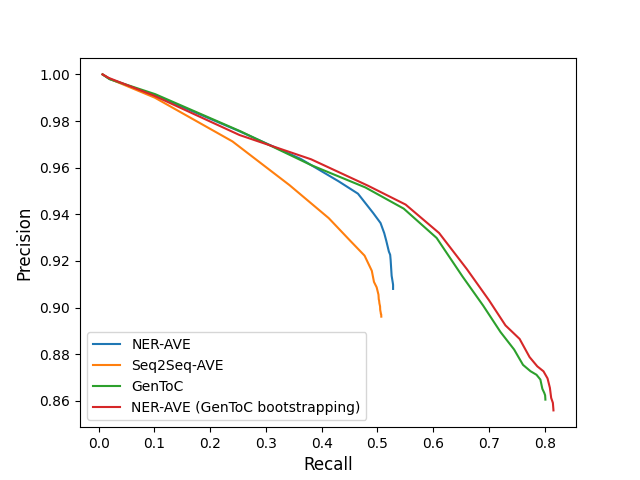}
    \caption{Precision-Recall curves show that \shortname{} and NER-AVE (\shortname{} bootstrapping) significantly outperform remaining models, NER-AVE and Seq2Seq-AVE.}
    \label{fig:pr-curve-main}
\end{figure}

To understand the trade-off between precision and recall across different systems, we evaluate the systems using a precision-recall curve.
We employ a common re-scoring model to compute the confidence level for all system extractions in this process. 
The model used for re-scoring is an independent Seq2Seq model, trained by using the \emph{product title} as the input and the `\textit{attribute: value}' string as the output, utilizing the same training data.
The model computes the confidence level associated with any given attribute-value pair by accumulating the log probabilities for every token present within the output string.
We found that the model provides us with better-calibrated scores for an attribute-value pair generated by any of the systems.
Using the results from the manually evaluated (2K) set, we applied uniformly spaced thresholds between 0 and 1 for the confidence values to create the Precision-Recall curve shown in \Cref{fig:pr-curve-main}.
Upon inspection, it was clear that \shortname{}'s performance significantly eclipsed that of baselines. %
Across most of the curve, \shortname{} achieves a higher precision for a given recall, demonstrating its power.
Further, the AUC in case of NER-AVE and Seq2Seq-AVE are 0.51 and 0.48 respectively, while that of \shortname{} is 0.76.
This substantiates the superior quality and efficacy of the \shortname{} system.

Additionally, the curve for NER-AVE trained with \shortname{} bootstrapped data closely mirrors that of \shortname{}, with an AUC of 0.77.
This further demonstrates the effectiveness of training with data bootstrapping from \shortname{}.

\subsection{Deployment Status and Impact}
\label{sec:status-and-impact}

Our system has been deployed on India's largest B2B e-commerce platform, IndiaMART, with \NumBuyers{} buyers and \NumSellers{} sellers.
It has been successfully integrated into the core product search functionality and has already served over \TotalRequests requests since deployment.
Specifically, it extracts attribute-value pairs from product titles in listings, and enables \textit{dynamic feature highlighting} of product features according to user search queries.
As an illustration of dynamic feature highlighting, when a user searches for \textit{``10 tier shoe rack''}, the system detects the attribute-value pair \textsc{Number of shelves}: \textit{10}. This information is then used to highlight the corresponding feature \textsc{Available number of shelves}: \textit{10} available for a listing with the product title \textit{``Steel Shoe Rack''}, in the search results.
This leads to enriched user search experience.

For the deployment version, we train \shortname{} using a dataset that is 3x larger, with a similar partially-labeled nature.
Further, we regenerate the training dataset using the trained \shortname{} model and subsequently train a faster NER-AVE model with this bootstrapped data, as detailed in \Cref{sec:bootstrapping}.
We use this \shortname{}-bootstrapped NER-AVE model for deployment.

\begin{table}[t]
\centering
\caption{Offline evaluation of previously deployed model and proposed model.}
\label{tab:deployment-offline-evaluation}
\begin{tabular}{@{}lrrr@{}}
\toprule
\textbf{System}                           & \textbf{Precision}    & \textbf{Recall}    & \textbf{F1-score} \\ \midrule
Rule-based (prior deployment)    & \textbf{91.9} & 70.9  & 80.0  \\
Proposed method                  & 89.5     & \textbf{91.1} & \textbf{90.3} \\ 
\bottomrule
\end{tabular}
\end{table}

We find significant improvements over the prior deployed system (which is based on rule-based non-neural system) for this task.
Rule-based systems that use regular-expression based techniques have their own limitations, such as struggling with negations (e.g., \textit{non-stretchable jeans}), handling common spelling variations (\textit{litre}, \textit{liter}, \textit{l}, or \textit{ltr}), making broad generalizations (both \textit{red} and \textit{raging red} refer to colors), and properly tagging attributes contextually (\textit{Galaxy} could refer to either a chocolate brand or a Samsung mobile model).

In an offline evaluation, based on a manual audit of search queries, we observe the results tabulated in \Cref{tab:deployment-offline-evaluation}.
It can be observed that the previously deployed rule-based system falls short of our proposed approach by more than 20\% in terms of recall, while being only marginally better by less than 2.5\% in precision.
This results in an overall difference of approximately 10\% in the F1-score between the two methods.
In online evaluation, we find that our system leads to a 9\% increase in queries with identified attributes that lead to dynamic feature highlighting.

\section{Conclusion and Future Work}

In this work, we introduce a new framework designed to effectively utilize incomplete training data, which is common in attribute-value extraction tasks, and provide a solution suitable for real-time deployment.
We achieve this by employing the novel \shortname{} model for attribute-value extraction, incorporating \textit{Markers} to learn from partially labeled data and \textit{Value Pruning} to avoid erroneous attribute tagging.
Moreover, by utilizing \shortname{}’s ability to enhance training data through bootstrapping, we can train faster NER models that are not designed to learn from partially labeled data, thereby bridging the gap between research and practical application.
Our system has been deployed on India's largest B2B e-commerce platform, IndiaMART, into the core product search functionality.
Future work could explore performing multiple stages of bootstrapping, handling near-redundancy in attributes, and uncovering novel attributes.

\begin{acks}
We would like to thank IndiaMART for providing us with the data for our research and helping us integrate with their current systems. We would like to thank KnowDis Data Annotator team for their timely help with the annotations required for this project and KnowDis Data Science members for giving valuable suggestions.
\end{acks}

\bibliographystyle{ACM-Reference-Format}
\bibliography{gentoc}


\begin{thebibliography}{28}


\ifx \showCODEN    \undefined \def \showCODEN     #1{\unskip}     \fi
\ifx \showDOI      \undefined \def \showDOI       #1{#1}\fi
\ifx \showISBNx    \undefined \def \showISBNx     #1{\unskip}     \fi
\ifx \showISBNxiii \undefined \def \showISBNxiii  #1{\unskip}     \fi
\ifx \showISSN     \undefined \def \showISSN      #1{\unskip}     \fi
\ifx \showLCCN     \undefined \def \showLCCN      #1{\unskip}     \fi
\ifx \shownote     \undefined \def \shownote      #1{#1}          \fi
\ifx \showarticletitle \undefined \def \showarticletitle #1{#1}   \fi
\ifx \showURL      \undefined \def \showURL       {\relax}        \fi
\providecommand\bibfield[2]{#2}
\providecommand\bibinfo[2]{#2}
\providecommand\natexlab[1]{#1}
\providecommand\showeprint[2][]{arXiv:#2}

\bibitem[Bing et~al\mbox{.}(2012)]%
        {bing2012unsupervised}
\bibfield{author}{\bibinfo{person}{Lidong Bing}, \bibinfo{person}{Tak-Lam Wong}, {and} \bibinfo{person}{Wai Lam}.} \bibinfo{year}{2012}\natexlab{}.
\newblock \showarticletitle{Unsupervised extraction of popular product attributes from web sites}. In \bibinfo{booktitle}{\emph{Information Retrieval Technology}}. Springer Berlin Heidelberg, \bibinfo{pages}{437--446}.
\newblock


\bibitem[{De Cao} et~al\mbox{.}(2021)]%
        {decao2021autoregressive}
\bibfield{author}{\bibinfo{person}{Nicola {De Cao}}, \bibinfo{person}{Gautier Izacard}, \bibinfo{person}{Sebastian Riedel}, {and} \bibinfo{person}{Fabio Petroni}.} \bibinfo{year}{2021}\natexlab{}.
\newblock \showarticletitle{Autoregressive Entity Retrieval}. In \bibinfo{booktitle}{\emph{9th International Conference on Learning Representations, {ICLR} 2021, Virtual Event, Austria, May 3-7, 2021}}. \bibinfo{publisher}{OpenReview.net}.
\newblock
\urldef\tempurl%
\url{https://openreview.net/forum?id=5k8F6UU39V}
\showURL{%
\tempurl}


\bibitem[Fleiss(1971)]%
        {Fleiss1971MeasuringNS}
\bibfield{author}{\bibinfo{person}{Joseph~L. Fleiss}.} \bibinfo{year}{1971}\natexlab{}.
\newblock \showarticletitle{Measuring nominal scale agreement among many raters.}
\newblock \bibinfo{journal}{\emph{Psychological Bulletin}}  \bibinfo{volume}{76} (\bibinfo{year}{1971}), \bibinfo{pages}{378--382}.
\newblock
\urldef\tempurl%
\url{https://api.semanticscholar.org/CorpusID:143544759}
\showURL{%
\tempurl}


\bibitem[Ghani et~al\mbox{.}(2006)]%
        {Ghani2006TextMF}
\bibfield{author}{\bibinfo{person}{Rayid Ghani}, \bibinfo{person}{Katharina Probst}, \bibinfo{person}{Yan Liu}, \bibinfo{person}{Marko Krema}, {and} \bibinfo{person}{Andrew~E. Fano}.} \bibinfo{year}{2006}\natexlab{}.
\newblock \showarticletitle{Text mining for product attribute extraction}.
\newblock \bibinfo{journal}{\emph{SIGKDD Explor.}}  \bibinfo{volume}{8} (\bibinfo{year}{2006}), \bibinfo{pages}{41--48}.
\newblock


\bibitem[Gopalakrishnan et~al\mbox{.}(2012)]%
        {Gopalakrishnan2012MatchingPT}
\bibfield{author}{\bibinfo{person}{Vishrawas Gopalakrishnan}, \bibinfo{person}{Suresh Iyengar}, \bibinfo{person}{Amit Madaan}, \bibinfo{person}{Rajeev Rastogi}, {and} \bibinfo{person}{Srinivasan~H. Sengamedu}.} \bibinfo{year}{2012}\natexlab{}.
\newblock \showarticletitle{Matching product titles using web-based enrichment}.
\newblock \bibinfo{journal}{\emph{Proceedings of the 21st ACM international conference on Information and knowledge management}} (\bibinfo{year}{2012}).
\newblock


\bibitem[He et~al\mbox{.}(2021)]%
        {debertav3}
\bibfield{author}{\bibinfo{person}{Pengcheng He}, \bibinfo{person}{Xiaodong Liu}, \bibinfo{person}{Jianfeng Gao}, {and} \bibinfo{person}{Weizhu Chen}.} \bibinfo{year}{2021}\natexlab{}.
\newblock \showarticletitle{DEBERTA: DECODING-ENHANCED BERT WITH DISENTANGLED ATTENTION}. In \bibinfo{booktitle}{\emph{International Conference on Learning Representations}}.
\newblock
\urldef\tempurl%
\url{https://openreview.net/forum?id=XPZIaotutsD}
\showURL{%
\tempurl}


\bibitem[Khandelwal et~al\mbox{.}(2023)]%
        {Khandelwal2023LargeSG}
\bibfield{author}{\bibinfo{person}{Anant Khandelwal}, \bibinfo{person}{Happy Mittal}, \bibinfo{person}{Shreyas~Sunil Kulkarni}, {and} \bibinfo{person}{Deepak~Kumar Gupta}.} \bibinfo{year}{2023}\natexlab{}.
\newblock \showarticletitle{Large Scale Generative Multimodal Attribute Extraction for E-commerce Attributes}.
\newblock \bibinfo{journal}{\emph{ArXiv}}  \bibinfo{volume}{abs/2306.00379} (\bibinfo{year}{2023}).
\newblock


\bibitem[Landis and Koch(1977)]%
        {Landis1977}
\bibfield{author}{\bibinfo{person}{J.~Richard Landis} {and} \bibinfo{person}{Gary~G. Koch}.} \bibinfo{year}{1977}\natexlab{}.
\newblock \showarticletitle{The Measurement of Observer Agreement for Categorical Data}.
\newblock \bibinfo{journal}{\emph{Biometrics}} \bibinfo{volume}{33}, \bibinfo{number}{1} (\bibinfo{year}{1977}), \bibinfo{pages}{159--174}.
\newblock
\showISSN{0006341X, 15410420}
\urldef\tempurl%
\url{http://www.jstor.org/stable/2529310}
\showURL{%
\tempurl}


\bibitem[Lewis et~al\mbox{.}(2019)]%
        {mbart}
\bibfield{author}{\bibinfo{person}{Mike Lewis}, \bibinfo{person}{Yinhan Liu}, \bibinfo{person}{Naman Goyal}, \bibinfo{person}{Marjan Ghazvininejad}, \bibinfo{person}{Abdelrahman Mohamed}, \bibinfo{person}{Omer Levy}, \bibinfo{person}{Veselin Stoyanov}, {and} \bibinfo{person}{Luke Zettlemoyer}.} \bibinfo{year}{2019}\natexlab{}.
\newblock \showarticletitle{{BART:} Denoising Sequence-to-Sequence Pre-training for Natural Language Generation, Translation, and Comprehension}.
\newblock \bibinfo{journal}{\emph{CoRR}}  \bibinfo{volume}{abs/1910.13461} (\bibinfo{year}{2019}).
\newblock
\showeprint[arXiv]{1910.13461}
\urldef\tempurl%
\url{http://arxiv.org/abs/1910.13461}
\showURL{%
\tempurl}


\bibitem[Nadeau and Sekine(2007)]%
        {Nadeau2007ASO}
\bibfield{author}{\bibinfo{person}{David Nadeau} {and} \bibinfo{person}{Satoshi Sekine}.} \bibinfo{year}{2007}\natexlab{}.
\newblock \showarticletitle{A survey of named entity recognition and classification}.
\newblock \bibinfo{journal}{\emph{Lingvisticae Investigationes}}  \bibinfo{volume}{30} (\bibinfo{year}{2007}), \bibinfo{pages}{3--26}.
\newblock


\bibitem[Probst et~al\mbox{.}(2007)]%
        {probst2007semi}
\bibfield{author}{\bibinfo{person}{Katharina Probst}, \bibinfo{person}{Rayid Ghani}, \bibinfo{person}{Marko Krema}, \bibinfo{person}{Andrew~E Fano}, {and} \bibinfo{person}{Yan Liu}.} \bibinfo{year}{2007}\natexlab{}.
\newblock \showarticletitle{Semi-supervised learning of attribute-value pairs from product descriptions}. In \bibinfo{booktitle}{\emph{Proceedings of the 20th International Joint Conference on Artificial Intelligence}}. Morgan Kaufmann Publishers Inc., \bibinfo{pages}{2838--2843}.
\newblock


\bibitem[Putthividhya and Hu(2011)]%
        {putthividhya2011bootstrapped}
\bibfield{author}{\bibinfo{person}{Duangmanee Putthividhya} {and} \bibinfo{person}{Junling Hu}.} \bibinfo{year}{2011}\natexlab{}.
\newblock \showarticletitle{Bootstrapped named entity recognition for product attribute extraction}. In \bibinfo{booktitle}{\emph{Proceedings of the 2011 Conference on Empirical Methods in Natural Language Processing}}. Association for Computational Linguistics, \bibinfo{pages}{1557--1567}.
\newblock


\bibitem[Rezk et~al\mbox{.}(2019)]%
        {Rezk2019AccuratePA}
\bibfield{author}{\bibinfo{person}{Mart{\'i}n Rezk}, \bibinfo{person}{Laura~Alonso Alemany}, \bibinfo{person}{Lasguido Nio}, {and} \bibinfo{person}{Ted Zhang}.} \bibinfo{year}{2019}\natexlab{}.
\newblock \showarticletitle{Accurate Product Attribute Extraction on the Field}.
\newblock \bibinfo{journal}{\emph{2019 IEEE 35th International Conference on Data Engineering (ICDE)}} (\bibinfo{year}{2019}), \bibinfo{pages}{1862--1873}.
\newblock


\bibitem[Ricatte and Crisostomi(2023)]%
        {ricatte2023avengr}
\bibfield{author}{\bibinfo{person}{Thomas Ricatte} {and} \bibinfo{person}{Donato Crisostomi}.} \bibinfo{year}{2023}\natexlab{}.
\newblock \showarticletitle{{AVEN}-{GR}: Attribute Value Extraction and Normalization using product {GR}aphs}. In \bibinfo{booktitle}{\emph{Proceedings of the 61st Annual Meeting of the Association for Computational Linguistics (Volume 5: Industry Track)}}, \bibfield{editor}{\bibinfo{person}{Sunayana Sitaram}, \bibinfo{person}{Beata Beigman~Klebanov}, {and} \bibinfo{person}{Jason~D Williams}} (Eds.). \bibinfo{publisher}{Association for Computational Linguistics}, \bibinfo{address}{Toronto, Canada}.
\newblock


\bibitem[Roy et~al\mbox{.}(2021)]%
        {Roy2021AttributeVG}
\bibfield{author}{\bibinfo{person}{Kalyani Roy}, \bibinfo{person}{Pawan Goyal}, {and} \bibinfo{person}{Manish Pandey}.} \bibinfo{year}{2021}\natexlab{}.
\newblock \showarticletitle{Attribute Value Generation from Product Title using Language Models}.
\newblock \bibinfo{journal}{\emph{Proceedings of The 4th Workshop on e-Commerce and NLP}} (\bibinfo{year}{2021}).
\newblock


\bibitem[Roy et~al\mbox{.}(2022)]%
        {Roy2022ExploringGM}
\bibfield{author}{\bibinfo{person}{Kalyani Roy}, \bibinfo{person}{Tapas Nayak}, {and} \bibinfo{person}{Pawan Goyal}.} \bibinfo{year}{2022}\natexlab{}.
\newblock \showarticletitle{Exploring Generative Models for Joint Attribute Value Extraction from Product Titles}.
\newblock \bibinfo{journal}{\emph{ArXiv}}  \bibinfo{volume}{abs/2208.07130} (\bibinfo{year}{2022}).
\newblock


\bibitem[Shinzato and Sekine(2013)]%
        {shinzato2013unsupervised}
\bibfield{author}{\bibinfo{person}{Keiji Shinzato} {and} \bibinfo{person}{Satoshi Sekine}.} \bibinfo{year}{2013}\natexlab{}.
\newblock \showarticletitle{Unsupervised extraction of attributes and their values from product description}. In \bibinfo{booktitle}{\emph{Proceedings of the Sixth International Joint Conference on Natural Language Processing}}. Asian Federation of Natural Language Processing, \bibinfo{pages}{1339--1347}.
\newblock


\bibitem[Shinzato et~al\mbox{.}(2022)]%
        {Shinzato2022SimpleAE}
\bibfield{author}{\bibinfo{person}{Keiji Shinzato}, \bibinfo{person}{Naoki Yoshinaga}, \bibinfo{person}{Yandi Xia}, {and} \bibinfo{person}{Wei-Te Chen}.} \bibinfo{year}{2022}\natexlab{}.
\newblock \showarticletitle{Simple and Effective Knowledge-Driven Query Expansion for QA-Based Product Attribute Extraction}.
\newblock \bibinfo{journal}{\emph{ArXiv}}  \bibinfo{volume}{abs/2206.14264} (\bibinfo{year}{2022}).
\newblock


\bibitem[Shinzato et~al\mbox{.}(2023)]%
        {Shinzato2023AUG}
\bibfield{author}{\bibinfo{person}{Keiji Shinzato}, \bibinfo{person}{Naoki Yoshinaga}, \bibinfo{person}{Yandi Xia}, {and} \bibinfo{person}{Wei-Te Chen}.} \bibinfo{year}{2023}\natexlab{}.
\newblock \showarticletitle{A Unified Generative Approach to Product Attribute-Value Identification}.
\newblock \bibinfo{journal}{\emph{ArXiv}}  \bibinfo{volume}{abs/2306.05605} (\bibinfo{year}{2023}).
\newblock


\bibitem[Shrimal et~al\mbox{.}(2022)]%
        {Shrimal2022NERMQMRCFN}
\bibfield{author}{\bibinfo{person}{Anubhav Shrimal}, \bibinfo{person}{Avi~Rajesh Jain}, \bibinfo{person}{Kartik Mehta}, {and} \bibinfo{person}{Promod Yenigalla}.} \bibinfo{year}{2022}\natexlab{}.
\newblock \showarticletitle{NER-MQMRC: Formulating Named Entity Recognition as Multi Question Machine Reading Comprehension}.
\newblock \bibinfo{journal}{\emph{ArXiv}}  \bibinfo{volume}{abs/2205.05904} (\bibinfo{year}{2022}).
\newblock


\bibitem[Wang et~al\mbox{.}(2023)]%
        {Wang2023MPKGACMP}
\bibfield{author}{\bibinfo{person}{K. Wang}, \bibinfo{person}{Jianzhi Shao}, \bibinfo{person}{T. Zhang}, \bibinfo{person}{Qijin Chen}, {and} \bibinfo{person}{Chengfu Huo}.} \bibinfo{year}{2023}\natexlab{}.
\newblock \showarticletitle{MPKGAC: Multimodal Product Attribute Completion in E-commerce}.
\newblock \bibinfo{journal}{\emph{Companion Proceedings of the ACM Web Conference 2023}} (\bibinfo{year}{2023}).
\newblock
\urldef\tempurl%
\url{https://api.semanticscholar.org/CorpusID:258377639}
\showURL{%
\tempurl}


\bibitem[Wang et~al\mbox{.}(2020)]%
        {Wang2020LearningTE}
\bibfield{author}{\bibinfo{person}{Qifan Wang}, \bibinfo{person}{Li Yang}, \bibinfo{person}{Bhargav Kanagal}, \bibinfo{person}{Sumit~K. Sanghai}, \bibinfo{person}{D. Sivakumar}, \bibinfo{person}{Bin Shu}, \bibinfo{person}{Zac Yu}, {and} \bibinfo{person}{Jonathan~L. Elsas}.} \bibinfo{year}{2020}\natexlab{}.
\newblock \showarticletitle{Learning to Extract Attribute Value from Product via Question Answering: A Multi-task Approach}.
\newblock \bibinfo{journal}{\emph{Proceedings of the 26th ACM SIGKDD International Conference on Knowledge Discovery \& Data Mining}} (\bibinfo{year}{2020}).
\newblock


\bibitem[Wong et~al\mbox{.}(2009)]%
        {Wong2009ScalableAE}
\bibfield{author}{\bibinfo{person}{Yuk~Wah Wong}, \bibinfo{person}{Dominic Widdows}, \bibinfo{person}{Tom Lokovic}, {and} \bibinfo{person}{Kamal Nigam}.} \bibinfo{year}{2009}\natexlab{}.
\newblock \showarticletitle{Scalable Attribute-Value Extraction from Semi-structured Text}.
\newblock \bibinfo{journal}{\emph{2009 IEEE International Conference on Data Mining Workshops}} (\bibinfo{year}{2009}), \bibinfo{pages}{302--307}.
\newblock


\bibitem[Xu et~al\mbox{.}(2019)]%
        {Xu2019ScalingUO}
\bibfield{author}{\bibinfo{person}{Huimin Xu}, \bibinfo{person}{Wenting Wang}, \bibinfo{person}{Xin Mao}, \bibinfo{person}{Xinyue Jiang}, {and} \bibinfo{person}{Man Lan}.} \bibinfo{year}{2019}\natexlab{}.
\newblock \showarticletitle{Scaling up Open Tagging from Tens to Thousands: Comprehension Empowered Attribute Value Extraction from Product Title}. In \bibinfo{booktitle}{\emph{ACL}}.
\newblock


\bibitem[Yang et~al\mbox{.}(2021)]%
        {Yang2021MAVEAP}
\bibfield{author}{\bibinfo{person}{Li Yang}, \bibinfo{person}{Qifan Wang}, \bibinfo{person}{Zac Yu}, \bibinfo{person}{Anand Kulkarni}, \bibinfo{person}{Sumit~K. Sanghai}, \bibinfo{person}{Bin Shu}, \bibinfo{person}{Jonathan~L. Elsas}, {and} \bibinfo{person}{Bhargav Kanagal}.} \bibinfo{year}{2021}\natexlab{}.
\newblock \showarticletitle{MAVE: A Product Dataset for Multi-source Attribute Value Extraction}.
\newblock \bibinfo{journal}{\emph{Proceedings of the Fifteenth ACM International Conference on Web Search and Data Mining}} (\bibinfo{year}{2021}).
\newblock


\bibitem[Zhang et~al\mbox{.}(2021)]%
        {Zhang2021QUEACOBT}
\bibfield{author}{\bibinfo{person}{Danqing Zhang}, \bibinfo{person}{Zheng Li}, \bibinfo{person}{Tianyu Cao}, \bibinfo{person}{Chen Luo}, \bibinfo{person}{Tony Wu}, \bibinfo{person}{Hanqing Lu}, \bibinfo{person}{Yiwei Song}, \bibinfo{person}{Bing Yin}, \bibinfo{person}{Tuo Zhao}, {and} \bibinfo{person}{Qiang Yang}.} \bibinfo{year}{2021}\natexlab{}.
\newblock \showarticletitle{QUEACO: Borrowing Treasures from Weakly-labeled Behavior Data for Query Attribute Value Extraction}.
\newblock \bibinfo{journal}{\emph{Proceedings of the 30th ACM International Conference on Information \& Knowledge Management}} (\bibinfo{year}{2021}).
\newblock
\urldef\tempurl%
\url{https://api.semanticscholar.org/CorpusID:237213565}
\showURL{%
\tempurl}


\bibitem[Zheng et~al\mbox{.}(2018)]%
        {Zheng2018OpenTagOA}
\bibfield{author}{\bibinfo{person}{Guineng Zheng}, \bibinfo{person}{Subhabrata Mukherjee}, \bibinfo{person}{Xin Dong}, {and} \bibinfo{person}{Feifei Li}.} \bibinfo{year}{2018}\natexlab{}.
\newblock \showarticletitle{OpenTag: Open Attribute Value Extraction from Product Profiles}.
\newblock \bibinfo{journal}{\emph{Proceedings of the 24th ACM SIGKDD International Conference on Knowledge Discovery \& Data Mining}} (\bibinfo{year}{2018}).
\newblock


\bibitem[Zhu et~al\mbox{.}(2020)]%
        {Zhu2020MultimodalJA}
\bibfield{author}{\bibinfo{person}{Tiangang Zhu}, \bibinfo{person}{Yue Wang}, \bibinfo{person}{Haoran Li}, \bibinfo{person}{Youzheng Wu}, \bibinfo{person}{Xiaodong He}, {and} \bibinfo{person}{Bowen Zhou}.} \bibinfo{year}{2020}\natexlab{}.
\newblock \showarticletitle{Multimodal Joint Attribute Prediction and Value Extraction for E-commerce Product}. In \bibinfo{booktitle}{\emph{Conference on Empirical Methods in Natural Language Processing}}.
\newblock


\end{thebibliography}

\clearpage
\appendix

\section{\shortname{} Algorithms}

\begin{algorithm}[H]
\caption{Training algorithm for \shortname{}}
\label{algo:training}
\begin{algorithmic}
\State $\textbf{Input:}$ Set of <Query, Incomplete attribute-value pairs>
\State $\textbf{Output:}$ Gen-AE and ToC-VE models
\State \textbf{Training Gen-AE Model:}
\For{<query, attribute-value pairs> in training set}
    \State \textit{Step-1:} Tag the words as \textit{marked} in the query if they are present in any of the attribute values.
    \State \textit{Step-2:} Train Gen-AE with marker embeddings added to the encoder's hidden states of the marked words.
    \State \textit{Step-3:} Gen-AE is trained to generate the attribute list in order of occurrence in the query.
\EndFor
\State
\State \textbf{Training ToC-VE Model:}
\For{each query in training set}
    \For{each attribute-value pair associated with the query}
        \State \textit{Step 1:} Concatenate both the attribute name and query with \textit{<sep>} delimiter.
        \State \textit{Step 2:} Label tokens as `YES' if they are present in the value for the attribute; otherwise `NO'.
        \State \textit{Step 3:} Apply Value Pruning to create examples with no values for a particular attribute.
        \State \textit{Step 4:} Train ToC-VE model for binary classification of token values (`YES'/`NO').
    \EndFor
\EndFor
\end{algorithmic}
\end{algorithm}

\begin{algorithm}[H]
\caption{Inference algorithm for \shortname{}}
\label{algo:inference}
\begin{algorithmic}
\State \textbf{Input:} query 
\State \textbf{Output:} List of attribute-value pairs 
\State \textit{Step 1:} Tag all words in the query as \textit{marked}.
\State \textit{Step 2:} Use the trained Gen-AE model to predict attributes for the query with marker embeddings added to encoder hidden states of the marked words.
\State \textit{Step 3:} For each predicted attribute from Gen-AE:
    \State \quad \textit{Step 3.1:} Concatenate the predicted attribute and query with \textit{<sep>} delimiter.
    \State \quad \textit{Step 3.2:} Use the trained ToC-VE model to classify each token as a value ('YES') or not ('NO').
    \State \quad \textit{Step 3.3:} Extract values for each attribute, excluding attributes with no associated values.
\end{algorithmic}
\end{algorithm}

\section{Performance on Long Product Titles}
\label{sec:performance-long}

To assess how well the models handle long-tail cases, we evaluate their performance on long product titles. Product titles consist of five words on average, with a standard deviation of two. Thus, we create a test set comprising \NumLongTestingSamples{} examples, each containing at least \ThreshLongTestingSamples{} words, for the evaluation process.
In \Cref{tab:performance-long-all}, we tabulate the scores obtained from manual evaluation, together with the \textit{tagged ratio} — that is, the proportion of words in the product title that have been marked with an attribute — for various models.

While Seq2Seq-AVE exhibits the highest precision at 95.3\%, and \shortname{} without VP shows a notable recall of 90.3\%, \shortname{}, stands out with the highest F1-score of 90.7\%. This underscores \shortname{}'s superior balance in the precision-recall trade-off, even in cases where the product titles are long.
It's noteworthy that when Markers are incorporated into the NER-AVE and Seq2Seq-AVE models, over 99\% of the words in product titles are linked to an attribute.
However, this high tagging ratio might not be ideal, particularly for lengthy queries, as not every word necessarily possesses a relevant attribute.
This is also reflected in the significant decrease in precision values for these models when Markers are used — a drop of over 22\% compared to their counterparts without Markers.
Single-stage models, which have to perform value labeling alongside attribute extraction, face this issue when Markers are employed, as they attempt to arbitrarily assign an attribute to every word. 
However, since we apply markers only to the first stage of the \shortname{} model, it does not face this problem as there is no direct one-to-one mapping between the predicted attributes and input words.
Finally, even the NER-AVE model trained with \shortname{} bootstrapping attains an F1-score that's comparable with \shortname{}, while having a faster response time.
This shows that the full potential of NER-AVE model is realized only with high-quality training data.

\begin{table}[ht]
\small
\centering
\caption{Results of manual evaluation for various models on a test set of 500 long product titles.}
\label{tab:performance-long-all}
\begin{tabular}{@{}lrrrr@{}}
\toprule
\textbf{Architecture} & \textbf{Precision} & \textbf{Recall} & \textbf{F1-score} & \textbf{\begin{tabular}[c]{@{}r@{}}Tagged\\ Ratio\end{tabular}} \\ \midrule
NER-AVE                                & 94.8  & 65.2  & 77.3                 & 0.464                                                                          \\
\ \ \ \ \ \ \ \ with Marker                                  & 72.4 & 90.3 & 80.4                       & \textbf{0.999}                                                                          \\ \midrule
Seq2Seq-AVE                                     & \textbf{95.3}  &  58.2 & 72.2                   & 0.403                                                                          \\
\ \ \ \ \ \ \ \ with Marker                                 & 71.7  & 71.6 &  71.6                  & 0.993                                                                          \\ \midrule
\shortname{}                             & 90.7 & 90.7 & \textbf{90.7}                      & 0.662                                                                          \\
\ \ \ \ \ \ \ \ without Marker                            & 94.5   & 58.7 &  72.4                  & 0.405                                                                          \\
\ \ \ \ \ \ \ \ without VP                                   & 87.1    & \textbf{93.0} & 89.9                  & 0.697                                                                          \\
\ \ \ \ \ \ \ \ without Marker \& VP                         & 94.2  & 59.0 & 72.6                    & 0.406                                                                          \\ \midrule
\begin{tabular}[c]{@{}l@{}}NER-AVE\\ (\shortname{} bootstrapping)\end{tabular}      & 90.5   & 88.4 & 89.4                  & 0.689                                                                          \\ 
\bottomrule
\end{tabular}
\end{table}

\section{Performance on Long-tail Attributes}
\label{sec:performance-long-tail-attributes}

\begin{figure}[h]
    \centering
    \includegraphics[width=\columnwidth]{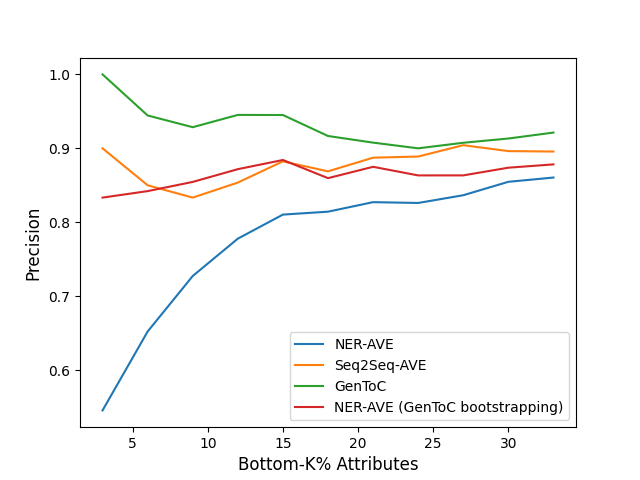}
    \caption{Performance of NER-AVE, Seq2Seq-AVE, \shortname{}, and \shortname{}-bootstrapped NER-AVE on long-tail attribute names. NER-AVE trained on original data shows poor performance on infrequent attribute names.}
    \label{fig:long-tail-attributes}
\end{figure}

To assess the efficacy of models in handling attributes with low occurrence rates, we measure their precision on attributes within the bottom 33\% by frequency in the training dataset.
These findings are illustrated in \Cref{fig:long-tail-attributes}.
Notably, the NER-AVE model, when trained on the original dataset, exhibits a markedly inferior performance, with a difference of up to 19\% compared to the \shortname{} model with the bottom-10\% of attributes.
This is likely due to the NER-AVE model treating each attribute name as an independent atomic label, which, combined with the limited training data for long-tail attributes, makes learning difficult during training.
Conversely, Seq2Seq-AVE and \shortname{} employ compositional encoding for attributes, enabling them to capture the semantic meaning of attribute names by considering the constituent words.
This approach allows them to handle long-tail and complex attributes more effectively than NER-AVE.
Additionally, it can be seen that NER-AVE with \shortname{} bootstrapping performs better than when trained with the original data. This improvement is likely due to the enrichment of attribute names provided by the \shortname{} model during the bootstrapping process.

\end{document}